\documentclass[letterpaper,10pt,journal,twoside]{IEEEtran}
\usepackage[letterpaper,margin=1in,bottom=1.05in]{geometry}
\usepackage{amsmath,amssymb}
\usepackage[T1,OT1]{fontenc}
\usepackage{graphicx}
\usepackage{multirow}
\usepackage{booktabs}
\usepackage[table,xcdraw]{xcolor}
\usepackage[nocompress]{cite}
\usepackage[hidelinks]{hyperref}

\newsavebox{\singlecoltablebox}
\newcommand{\singlecoltable}[3]{%
  \setlength{\tabcolsep}{0pt}%
  \sbox{\singlecoltablebox}{\begin{tabular}{#1}#3\end{tabular}}%
  \setlength{\tabcolsep}{\dimexpr(\columnwidth-\wd\singlecoltablebox)/#2\relax}%
  \begin{tabular}{#1}#3\end{tabular}%
}

\title{Shape Distribution Matters: Shape-Specific Mixture-of-Experts for Amodal Segmentation Under Diverse Occlusions}
\author{Zhixuan~Li, Yujia~Liu, Chen~Hui, Jeonghaeng~Lee, Sanghoon~Lee,~\IEEEmembership{Fellow,~IEEE,} and~Weisi~Lin,~\IEEEmembership{Fellow,~IEEE}%
\thanks{This work has been submitted to the IEEE for possible publication. Copyright may be transferred without notice, after which this version may no longer be accessible.}%
\thanks{This research was supported by the Ministry of Education of Singapore Tier3 grant MOET32022-0006. This work was supported in part by Culture, Sports and Tourism R\&D Program through Korea Creative Content Agency grant funded by the Ministry of Culture, Sports and Tourism in 2024 under Grant RS-2024-00398413. This work was supported by the National Natural Science Foundation of China under grant 62601410, Natural Science Foundation of Jiangsu under grant BK20250731, the Startup Foundation for Introducing Talent of Nanjing University of Information Science and Technology under grant 2025r029. This work was supported in part by the National Natural Science Foundation of China (Grant No. 62506009), the China National Postdoctoral Program for Innovative Talents (Grant No. BX2026295), and the China Postdoctoral Science Foundation (Grant No. 2026M791599). \textit{(Corresponding author: Weisi Lin.)}}%
\thanks{Zhixuan Li and Weisi Lin are with the College of Computing and Data Science, Nanyang Technological University, Singapore 639798 (e-mail: \href{mailto:zhixuanli520@gmail.com}{zhixuanli520@gmail.com}; \href{mailto:WSLin@ntu.edu.sg}{WSLin@ntu.edu.sg}).}%
\thanks{Yujia Liu is with the School of Computer Science, the National Key Laboratory for Multimedia Information Processing, and the National Engineering Research Center of Visual Technology, Peking University, Beijing 100871, China (e-mail: \href{mailto:yujia_liu@pku.edu.cn}{{\fontencoding{T1}\selectfont yujia\_liu@pku.edu.cn}}).}%
\thanks{Chen Hui is with the School of Artificial Intelligence, Nanjing University of Information Science and Technology, Nanjing 210044, China (e-mail: \href{mailto:chui@nuist.edu.cn}{chui@nuist.edu.cn}).}%
\thanks{Jeonghaeng Lee and Sanghoon Lee are with the Department of Electrical and Electronic Engineering, Yonsei University, Seoul 03722, Republic of Korea (e-mail: \href{mailto:leedoright@yonsei.ac.kr}{leedoright@yonsei.ac.kr}; \href{mailto:slee@yonsei.ac.kr}{slee@yonsei.ac.kr}).}}

\begin{document}
\bstctlcite{IEEEexample:BSTcontrol}
\maketitle

\begin{abstract}
Amodal segmentation aims to predict complete object masks covering both visible and occluded regions, serving as a critical foundation for robust multimedia scene understanding and interactive media editing. This task poses significant challenges in complex visual media due to severe occlusions and extreme shape variation, ranging from rigid furniture to highly deformable clothing. Existing one-size-fits-all approaches struggle to reason about diverse amodal shapes in multimedia content due to limited representation capacity. While a mixture-of-experts framework is a natural solution, naively applying it without considering the underlying shape distribution leads to mismatched routing and redundant experts. To address these issues, we introduce ShapeMoE, a shape-specific sparse mixture-of-experts framework designed for complex visual scene parsing. The key idea is to learn a latent shape distribution space and dynamically route each object to a lightweight expert tailored to its geometric characteristics. Specifically, ShapeMoE encodes each object into a compact Gaussian embedding that captures key shape features. A Shape-Aware Sparse Router then maps the object to the most suitable expert, enabling precise shape-aware routing. Each expert is lightweight and specializes in recovering occluded regions for specific shape patterns. ShapeMoE is designed for shape-specific expert specialization while maintaining computational efficiency. Extensive experiments on three challenging benchmarks demonstrate that ShapeMoE consistently outperforms state-of-the-art methods. Beyond advancing fundamental visual reasoning, ShapeMoE provides a robust and efficient visual prior for downstream multimedia applications, particularly in occlusion-heavy immersive environments.
\end{abstract}


\begin{figure*}[!t]
    \centering
    \includegraphics[width=\linewidth]{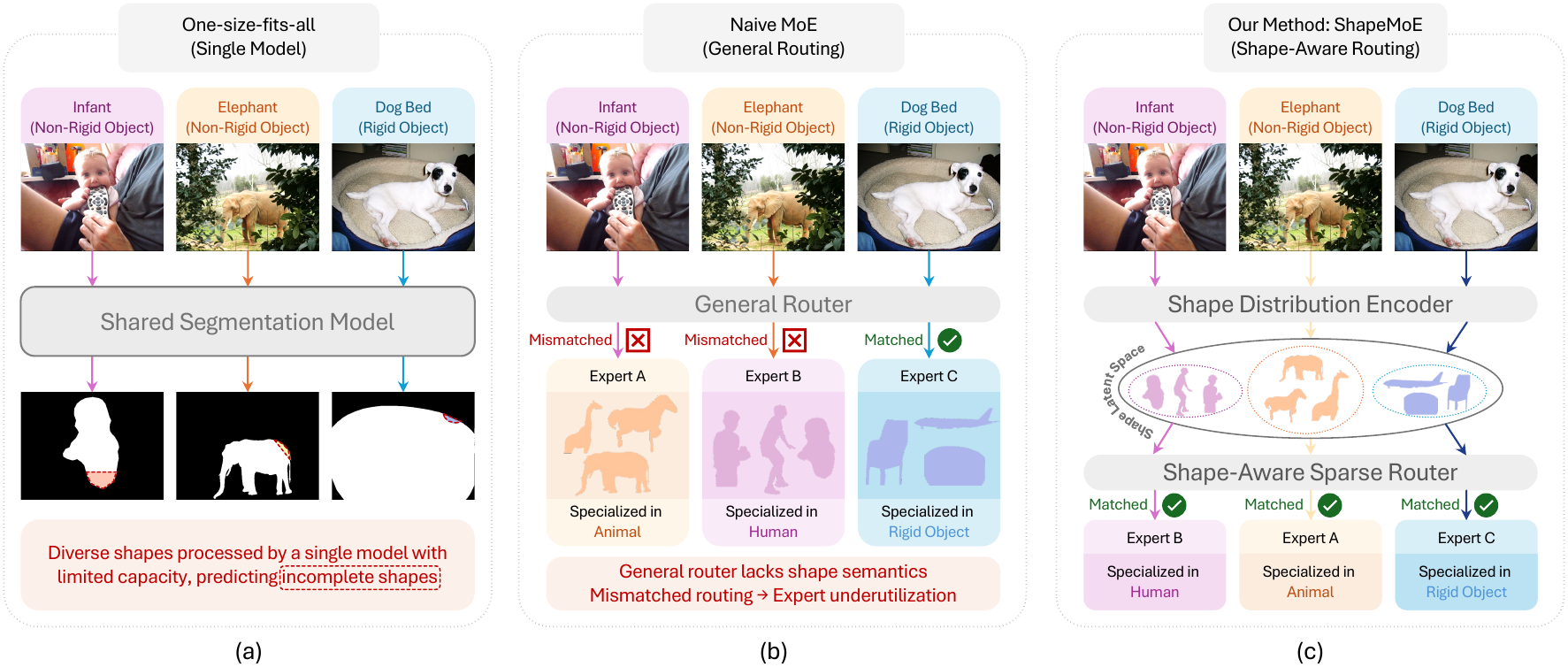}
    \caption{Motivation and Comparison of Routing Strategies. (a) One-size-fits-all models treat all shape types equally, often producing incomplete predictions under occlusion. 
    (b) Naive MoE approaches rely on softmax-based routing without modeling shape distributions, leading to a mismatch between samples and experts. 
    (c) Our ShapeMoE framework encodes each shape as a Gaussian distribution in a latent space, enabling shape-aware sparse routing to specialized experts and improving segmentation of diverse amodal shapes. Best viewed in color.
}
    \label{fig:introduction}
\end{figure*}

\section{Introduction}

Amodal instance segmentation~\cite{li2016amodal} aims to predict complete object masks, including occluded regions, serving as a foundational task for comprehensive multimedia scene understanding. A key challenge in processing diverse visual media lies in the wide variation of complete object shapes, making accurate segmentation under occlusion particularly difficult. Specifically, within complex multimedia content, some objects exhibit rigid and regular geometries, while others present highly deformable and irregular structures. Such high variation in shape introduces significant complexity to the visual learning process. 
Amodal segmentation has broad applicability in modern multimedia systems~\cite{lee2023feasibility,follmann2019learning,qi2019amodal}, especially in immersive multimedia (e.g., AR/VR/XR)~\cite{li2022real,yan2022augmented,casas2024exploring} and interactive content editing~\cite{kong2024omg,meng20223d,pan2023drag}. In these scenarios, robustly reasoning about occluded structures is critical for realistic 3D scene reconstruction, seamless object manipulation, and providing robust visual priors for multimodal foundation models.

Existing approaches~\cite{tai2025segment,zhan2024amodal,ozguroglu2024pix2gestalt,gao2023coarse,xiao2021amodal,zhan2020self} typically rely on a single model to predict amodal masks across all object categories and shapes. However, this one-size-fits-all strategy often leads to generic predictions due to limited model capacity, either failing to predict the precise geometry of occluded regions or generating implausible completions, making it difficult to handle complex occlusions and diverse object shapes. 

A natural solution to this challenge is to leverage the Mixture of Experts (MoE)~\cite{shazeer2017outrageously,riquelme2021scaling} framework, which utilizes multiple specialized experts to better capture the diverse shape variations of target objects. 
However, routers in existing MoE frameworks often rely on simple softmax-based gating networks that lack an understanding of fine-grained shape semantics. This limitation results in suboptimal routing, where samples are assigned to inappropriate experts specialized in certain shape patterns. Consequently, such mismatches between samples and experts lead to limited performance in the amodal mask prediction.

Moreover, the expert design in the existing MoE-based method for amodal instance segmentation, such as SAMBA~\cite{liu2025towards}, focuses solely on varying occlusion ratios within certain shapes, without explicitly considering the rich diversity of shape variations. 
For example, objects with distinct shapes (e.g., a dog vs. a car) can share the same occlusion ratio. While such significant structural disparities naturally demand different experts for amodal segmentation, SAMBA often assigns them to the same expert by relying solely on the occlusion ratio as its routing clue, which is highly coarse-grained compared to detailed shape representations.

In this paper, we propose ShapeMoE, a novel shape-aware sparse mixture of experts framework tailored for amodal instance segmentation, enabling flexible and efficient learning over highly diverse occluded shapes. The core idea is to model the shape characteristics of each object and route them to the most suitable expert for amodal segmentation. 
Specifically, our method consists of three key designs.
First, we design a shape modeling approach to capture the shape distribution of each object. Specifically, we generate shape-specific embeddings from its visible mask, which serve as input prompts for locating the corresponding target object. 
Assuming that object shapes follow a Gaussian distribution, we estimate the corresponding Gaussian parameters from the shape embeddings to effectively characterize the underlying shape distribution in the latent space.
This modeling enables a compact representation of each mask’s geometric identity, effectively capturing shape variations in a latent shape space, while facilitating clear discrimination between different shape patterns. 

Next, based on the predicted shape distribution, we present a gating network named Shape-Aware Sparse Router, which routes each sample according to its shape characteristics. Unlike conventional gating mechanisms that compute softmax weights directly from pooled features without explicit shape modeling, our router takes the predicted Gaussian parameters, which characterize the shape distribution, as input and determines the most appropriate expert, ensuring that each sample is processed by an expert specialized in its corresponding shape pattern. Moreover, by leveraging the sparse MoE mechanism~\cite{shazeer2017outrageously}, our method achieves high model capacity with various experts while maintaining computational efficiency, as only one expert is activated per sample. Furthermore, our design offers clear interpretability by enabling the statistical analysis and visualization of the shape patterns handled by each expert, revealing their specialization in distinct geometric structures.
Crucially, even for unseen objects, their shape embeddings can be projected into the same latent distributional space, enabling the router to effectively assign them to the most suitable expert and allowing the model to generalize beyond the shapes observed during training. 

Finally, we introduce a carefully designed shape-specialized segmentation expert to effectively handle diverse occlusion patterns in object shapes.
While directly replicating the entire mask decoder architecture, such as that in SAM, is a straightforward way to construct expert branches, it results in substantial computational overhead and redundant capacity.
To address this, we first analyze the decoder architecture to identify the components most critical for shape-specific reasoning. Based on this analysis, we selectively convert the compact, shape-sensitive components into multiple expert branches, introducing only marginal additional parameters. Each sample is then routed to the most appropriate expert via our Shape-Aware Sparse Router. This design allows each expert to focus on distinct amodal shape patterns while sharing common visual features, thereby enhancing the model’s capacity to reason about occlusion and capture intrinsic shape diversity.

To summarize, our contributions are as follows. 
First, we propose ShapeMoE, a sparse mixture-of-experts framework that, to the best of our knowledge, is the first to explicitly model shape diversity and dynamically route each instance to a specialized expert, alleviating the limitations of one-size-fits-all amodal segmentation models.
Second, we design a Gaussian-guided routing mechanism that enables interpretable, distribution-based expert assignment within a learned shape embedding space, while naturally generalizing to unseen shapes.
Third, we convert lightweight, shape-specific decoder components into multiple expert branches, enabling focused and efficient shape-aware learning with minimal additional cost while improving occlusion handling and capturing diverse object shapes.
Finally, we conduct comprehensive experiments on challenging amodal segmentation datasets, including COCOA-cls, KINS, and D2SA, demonstrating consistent improvements in amodal segmentation, especially the accuracy of occluded regions over state-of-the-art methods.

\section{Related Work}
\subsection{Amodal Segmentation}

Amodal Segmentation~\cite{li2016amodal,zhu2017semantic} comprises predicting the visible and occluded areas to obtain the complete shape of the target object. Most existing approaches~\cite{xiao2021amodal,sun2022amodal,li20222d,chen2023amodal,gao2023coarse,li2023gin,tran2024shapeformer} explore the usage of shape prior knowledge to enhance the prediction of amodal masks, while some approaches utilize specific designs such as semantics-aware distance map~\cite{zhang2019learning}, occlusion relationship construction~\cite{zhan2020self,back2022unseen,zheng2021visiting}, boundary uncertainty~\cite{nguyen2021weakly}, occlusion condition modeling~\cite{li2023oaformer} and multi-view relations learning~\cite{li2023muva,zhou2023few}. 

Recently, with the advancement of large language models (LLMs) and foundation models, several approaches have begun exploring the amodal segmentation task through these powerful paradigms. For example, AURA~\cite{li2025aura} presents a novel method and dataset that explores predicting visible and amodal masks by reasoning the implicit intention in the user's textual input. Zero-shot methods~\cite{liu2025towards,ao2025open} utilize off-the-shelf foundation models to directly predict amodal masks on unseen objects effectively. Moreover, SDAmodal~\cite{zhan2024amodal} reuses the knowledge of the Stable Diffusion model to generate amodal masks precisely, while SAMEO~\cite{tai2025segment} fine-tunes EfficientSAM~\cite{xiong2024efficientsam} on a synthetic dataset for generalizable amodal segmentation. 

Although these methods have achieved notable performance, all of these methods are designed to comprehend the occluded shape and predict amodal masks based on a single model with limited model capacity. To enhance the model learning capacity while keeping efficiency, a novel shape-aware mixture-of-experts framework named ShapeMoE is designed with a sparse paradigm. Based on our ShapeMoE, shape distributions are learned and dynamically assigned with shape-specific experts for predicting amodal masks adaptively.

\subsection{Mixture-of-Expert Mechanism}

Mixture of Experts (MoE) architectures have emerged as a compelling solution to scale model capacity while maintaining computational efficiency. Many approaches~\cite{shazeer2017outrageously,riquelme2021scaling} have demonstrated how sparse routing can activate only a subset of experts per input, significantly reducing computation without sacrificing performance. The integration of MoE with transformer-based architectures~\cite{lepikhin2020gshard,fedus2022switch} has further improved this paradigm, enabling large models with billions of parameters while keeping inference efficient. Moreover, recent lightweight transformers~\cite{zhou2025care} highlight the need to optimize computational overhead in complex tasks.

However, although directly integrating MoE with amodal segmentation methods can improve the capacity for shape learning, shape characteristics of occluded objects have not been explored in existing MoE frameworks. SAMBA~\cite{liu2025towards} relies on the object's occlusion ratio as the routing criterion, which provides a coarse-grained and inherently ambiguous signal for shape representations. Such a scalar metric severely lacks the fine-grained feature discrimination that has proven crucial in advanced prototype-based representation learning~\cite{zhou2022hierarchical}.

Considering there are no existing approaches exploring the MoE for shape modeling in the amodal segmentation task at present, we propose ShapeMoE to be aware of the shape distribution of each object to effectively assign objects to shape-specific experts for high-quality amodal segmentation. 
Unlike SAMBA, which uses the occlusion ratio as its routing criterion, our shape-specific design inherently provides better interpretability.

\begin{figure*}[ht]
    \centering
    \includegraphics[width=\linewidth]{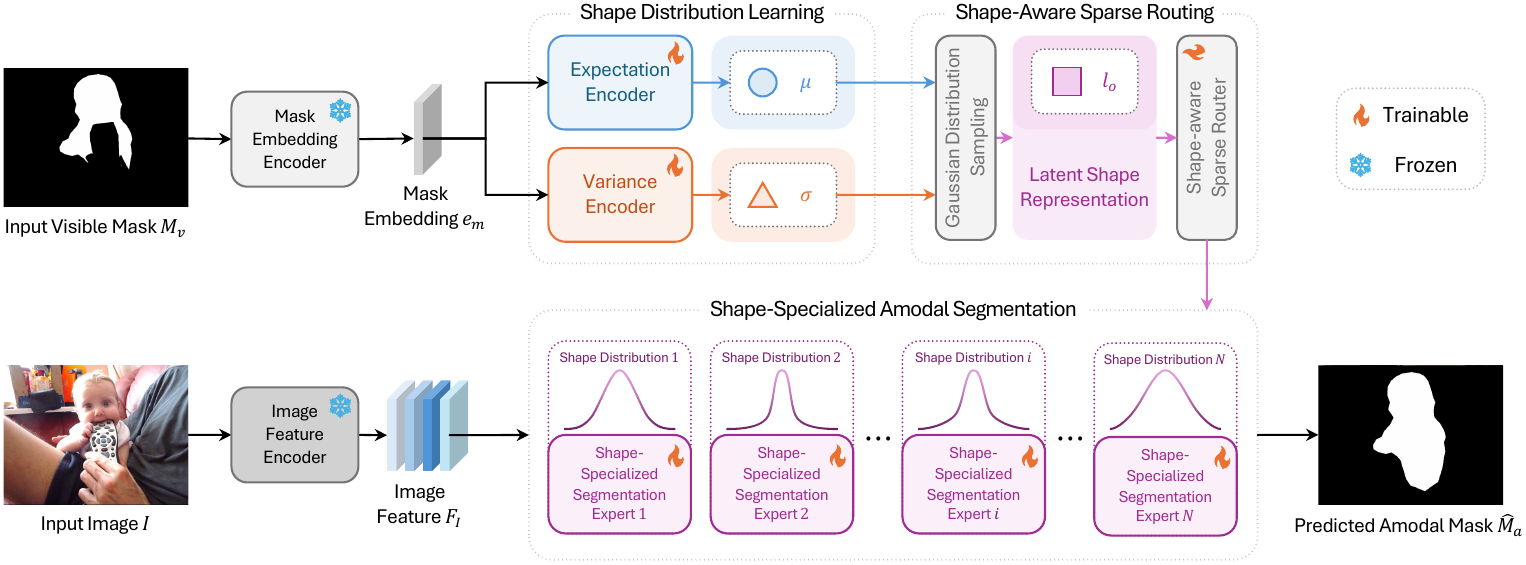}
    \caption{Given an input image and a visible mask, the default single-Gaussian ShapeMoE performs amodal segmentation through the following stages. (1) The image is encoded by the image feature encoder, while the visible mask is embedded into a shape-aware mask embedding. (2) The Shape Distribution Encoder predicts the Gaussian parameters that characterize the object’s shape distribution in a learned latent space. (3) A latent shape representation is sampled, and the Shape-Aware Sparse Router computes expert selection scores to route each instance to the most appropriate expert. (4) The selected expert, specialized in specific shape patterns, predicts the final high-quality amodal segmentation mask. The GMM extension is described in Section~\ref{sec:gmm_extension}. Best viewed in color.}
    \label{fig:overall_architecture}
\end{figure*}

\section{Methodology}


\subsection{Task Definition}

Amodal segmentation task takes an input image $I$ as input, and optionally takes prompts like visible mask $M_v$ for locating the target object. The output of a model is the amodal mask $\hat{M}_a$ of the target object, covering the visible and occluded regions.

\subsection{Overall Architecture}

In this paper, we present the designed ShapeMoE, a shape-aware sparse MoE framework for amodal segmentation. As shown in Fig.~\ref{fig:overall_architecture}, the overall architecture of ShapeMoE consists of four stages. (1) The input image $I$ is fed into the Image Feature Encoder for extracting the general image feature $F_I$, while the input visible mask $M_v$ is embedded by the Mask Embedding Encoder $\mathcal{E}_{M}$ for obtaining the mask embedding $e_m$, which can be regarded as a feature pointing to the target object. (2) The Shape Distribution Encoder $\mathcal{E}_{S}$ takes the mask embedding $e_m$, which contains the shape description of the object's visible region as input, predicting the mean $\mu$ and raw scale $\sigma$ that parameterize a Gaussian distribution representing the shape distribution of the processed object. (3) The Shape-Aware Sparse Router $\mathcal{R}$ then takes the predicted mean $\mu$ and raw scale $\sigma$ as input and samples a latent shape representation $l_o$ from the learned distribution. This representation is subsequently used to compute the expert selection scores $s$, enabling the assignment of the most suitable expert for each specific shape pattern. (4) Finally, the shape-specialized segmentation expert $\mathcal{D}$ selected by the router is employed to predict the final segmentation mask, leveraging its specialized knowledge of the assigned shape pattern.

\begin{figure}[tbp]
    \centering
    \includegraphics[width=\linewidth]{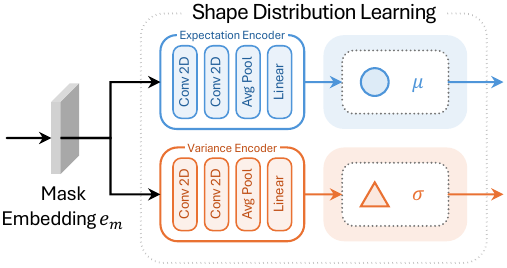}
    \caption{Architecture of the Shape Distribution Encoder.}
    \label{fig:shape_modeling_architecture}
\end{figure}

\subsection{Shape Distribution Learning}

To model the shape distribution of the target object, we design a Shape Distribution Encoder $\mathcal{E}_{S}$ to learn a probabilistic shape distribution from the mask embedding $e_m$ predicted by the Mask Embedding Encoder $\mathcal{E}_{M}$. As shown in Fig.~\ref{fig:shape_modeling_architecture}, the object's amodal shape distribution is assumed to follow a Gaussian distribution $\mathcal{N}(\mu, \operatorname{diag}(\operatorname{Softplus}(\sigma)^2))$. Based on this, two separate encoders, including Expectation Encoder $\mathcal{E}_{\mu}$ and Variance Encoder $\mathcal{E}_{\sigma}$ contained in the $\mathcal{E}_{S}$, are employed to predict the mean and raw scale, respectively:

\begin{equation}
\mu = \mathcal{E}_{\mu}(e_m), \quad \sigma = \mathcal{E}_{\sigma}(e_m),
\end{equation}

\noindent where $\mu, \sigma \in \mathbb{R}^{d}$ and $d$ is the latent dimension. The mean $\mu$ represents the center of the shape distribution. The raw scale $\sigma$ is transformed element-wise by Softplus to obtain the standard deviation, which captures shape variability or ambiguity. Here, $\operatorname{diag}(\cdot)$ constructs a diagonal covariance matrix from the element-wise squared standard deviations.

The mean $\mu$ and the standard deviation $\operatorname{Softplus}(\sigma)$ provide a probabilistic representation of shape characteristics, which can be leveraged by the router to effectively guide the expert selection.

\subsection{Shape-Aware Sparse Routing}

We present the Shape-Aware Sparse Router $\mathcal{R}$ to dynamically match between the mask embedding $e_m$ and the corresponding most suitable experts that specialize in the shape pattern. Specifically, the latent shape representation $l_o \in \mathbb{R}^{d}$ capturing the shape pattern of the target object is first sampled from the learned Gaussian distribution $\mathcal{N}(\mu, \operatorname{diag}(\operatorname{Softplus}(\sigma)^2))$ with the differentiable reparameterization method inspired by \cite{qiu2024duet}. The sampling process can be formulated as follows:

\begin{equation}
    l_o = \mu + \operatorname{Softplus}(\sigma) \odot \eta, \quad \eta \sim \mathcal{N}(\mathbf{0}, I_d)
\end{equation}

\noindent where $\odot$ denotes element-wise multiplication, and $\eta \in \mathbb{R}^{d}$ is standard Gaussian noise with independent components. Here, $I_d$ is the $d \times d$ identity matrix. Softplus ensures strictly positive standard deviations while preserving differentiability.

Next, the latent shape representation $l_o$ is utilized to predict the score $s \in \mathbb{R}^N$ for $N$ experts selection as follows:

\begin{equation}
    s = W l_o,
\end{equation}

\noindent where $W \in \mathbb{R}^{N \times d}$ is a trainable matrix that transforms the latent shape representation $l_o$  into expert scores $s$, representing the affinity between the sample and experts. During training, $W$ is optimized via backpropagation, so the model can learn how to route each latent shape representation $l_o$ to the most appropriate experts.

Finally, we compute $\tilde{\pi}=\operatorname{Softmax}(s)$ over all experts and retain the indices $\mathcal{I}_k$ of its top-$k$ entries. The sparse routing weights are defined as follows:
\begin{equation}
\pi_i =
\begin{cases}
\displaystyle\frac{\tilde{\pi}_i}{\sum_{j\in\mathcal{I}_k}\tilde{\pi}_j+\epsilon_g}, & i\in\mathcal{I}_k,\\
0, & \text{otherwise},
\end{cases}
\end{equation}
\noindent where $\pi\in\mathbb{R}^{N}$ and $\epsilon_g=10^{-6}$. In the default $N=4$, $k=1$ setting, the selected weight $\tilde{\pi}_i/(\tilde{\pi}_i+\epsilon_g)$ depends on the routing scores, retaining a nonzero gate derivative.

It is worth noting that only top-$k$ selected experts are assigned samples for segmentation, while unselected experts remain inactive, enabling increased model capacity for shape diversity without sacrificing efficiency.

\subsection{Shape-Specialized Segmentation Expert}

To enable high-quality amodal mask prediction, we introduce a shape-specialized segmentation expert. Given the strong generalization ability of the Segment Anything Model (SAM)~\cite{kirillov2023segment}, we adopt it as the base framework for implementing our expert design. SAM comprises an Image Feature Encoder for extracting general image features $F_I$, a Mask Embedding Encoder for embedding the input prompt, and a mask decoder that predicts segmentation masks based on the image feature and the embedded prompt. 

A naive implementation would duplicate the entire mask decoder $N$ times to instantiate multiple experts. However, this approach is inefficient and suboptimal, as not all components within the decoder contribute equally to shape prediction. Specifically, the two-way Transformer, the first stage of the decoder, is computationally expensive and primarily focuses on generating refined image-level features $F$ from the Image Feature Encoder output $F_I$, without directly contributing to mask generation. In contrast, the second stage, the hyper-network, is lightweight yet directly responsible for producing the mask weights $w$, which are subsequently multiplied with $F$ to generate the final masks. Therefore, this component is explicitly tied to shape reasoning and prediction.

Motivated by this analysis, we take the hyper-network as the core shape-specialized module and replicate it across $N$ expert branches with only marginal parameter overhead. This targeted design allows each expert to specialize in distinct shape patterns, guided by our Shape-Aware Sparse Router that dynamically assigns each sample to the most appropriate expert for predicting the amodal mask $\hat{M}_a$. As a result, our architecture achieves efficient and precise amodal segmentation through focused expert specialization.

\subsection{Loss Functions}

The overall loss function $\mathcal{L}$ comprises two components: one for supervising the predicted amodal segmentation masks and the other for balancing expert utilization. Specifically, the cross-entropy loss $\mathcal{L}_{CE}$ is used to supervise the amodal mask prediction, while a balancing loss $\mathcal{L}_{bal}$ based on the squared coefficient of variation ($CV^2$)~\cite{shazeer2017outrageously} is employed to encourage balanced expert selection. The overall loss function is defined as:

\begin{equation}
    \mathcal{L} = \mathcal{L}_{CE}(\hat{M}_a, M_a) + \mathcal{L}_{bal},
\end{equation}

\noindent where $M_a$ denotes the ground-truth amodal mask. Following DUET~\cite{qiu2024duet}, $\mathcal{L}_{bal}$ sums the squared coefficients of variation of expert importance and differentiable load estimates, each aggregated over the training batch. These load estimates are computed under noisy gating, allowing gradients to update the router even with top-1 expert selection.

\subsection{GMM Extension}
\label{sec:gmm_extension}

The preceding single-Gaussian model remains our default method. We additionally explore a Gaussian mixture model (GMM) for matching latent shapes to experts. The input-conditioned sampling distribution remains a single Gaussian, while the GMM introduces $N=4$ shared, learnable prototypes in the latent space. Component $j$ has center $m_j\in\mathbb{R}^{d}$, diagonal covariance $\Sigma_j=\operatorname{diag}(\tau_j^2)$, and prior $\beta_j$. We parameterize $\tau_j=\operatorname{Softplus}(\rho_j)+\delta$ and $\beta=\operatorname{Softmax}(b)$, where $\delta>0$ is a small constant and $\rho_j$ and $b$ are learnable. These prototypes replace the linear scoring matrix $W$ with Gaussian matching scores and component responsibilities:
\begin{equation}
\begin{aligned}
    s_j^{\mathrm{GMM}}(l_o) &= \log\beta_j+\log\mathcal{N}(l_o;m_j,\Sigma_j),\\
    r(l_o) &= \operatorname{Softmax}(s^{\mathrm{GMM}}(l_o)).
\end{aligned}
\end{equation}

Component $j$ is fixed to expert $j$, whose sigmoid mask probabilities are denoted by $P_j$. All four expert outputs are fused during both training and inference:
\begin{equation}
    P(l_o)=\sum_{j=1}^{N}r_j(l_o)P_j.
\end{equation}
Training uses the sampled $l_o$, whereas inference substitutes the encoded mean $\mu$. The image encoder and two-way Transformer are shared and evaluated once, but all four lightweight expert heads execute. Thus, this extension uses dense weighted fusion rather than the default model's sparse expert selection.

Each sample should favor experts suited to its shape while avoiding persistent concentration of weights on one expert across samples. Inspired by the balanced assignments in SwAV~\cite{caron2020unsupervised}, we score the current latent vector together with detached historical vectors using the current Gaussian parameters. For $B_{\mathrm{eff}}$ vectors, the score matrix has entries $S_{ij}=s_j^{\mathrm{GMM}}(l_{o,i})$. Log-domain Sinkhorn normalization of $S/\epsilon_{\mathrm{SK}}$ produces a nonnegative target matrix $A$, targeting row sums of $1$ and column sums of $B_{\mathrm{eff}}/N$. The current row supplies the detached target $a_j=\operatorname{stopgrad}(A_{\mathrm{current},j})$. It supervises temperature-scaled predictions $\tilde r=\operatorname{Softmax}(s^{\mathrm{GMM}}(l_o)/T)$ through the following loss:
\begin{equation}
\begin{aligned}
    \mathcal{L}_{\mathrm{bal}}^{\mathrm{GMM}} &=-\sum_{j=1}^{N}a_j\log\tilde r_j,\\
    \mathcal{L}_{\mathrm{GMM}} &=\operatorname{BCE}(P(l_o),M_a)+\mathcal{L}_{\mathrm{bal}}^{\mathrm{GMM}}.
\end{aligned}
\end{equation}
Only the current sample contributes to this cross-entropy term. The fusion weights remain $r$, and inference requires neither the queue nor Sinkhorn normalization.

\section{Experiments}

\begin{table*}[!t]
\centering
\caption{
    Comparison on amodal segmentation performance on three datasets, including COCOA-cls, D2SA, and KINS. SG and GMM denote the single-Gaussian baseline and its Gaussian mixture model extension, respectively. $\dag$ denotes a zero-shot method which was pretrained on a large-scale amodal segmentation dataset~\cite{ozguroglu2024pix2gestalt}. Larger metric values of mIoU$_{\text{full}}$ and mIoU$_{\text{occ}}$ indicate better performance. Best performance is in bold, and second-best performance is underlined for each metric. A dash indicates an unreported result 
}
\begin{tabular}{lccccccc}
\toprule
\multicolumn{1}{l}{\multirow{2}{*}{Methods}} &
  \multicolumn{1}{c}{\multirow{2}{*}{Venue}} &
  \multicolumn{2}{c}{COCOA-cls   (Val)} &
  \multicolumn{2}{c}{D2SA   (Val)} &
  \multicolumn{2}{c}{KINS   (Test)} \\ \cmidrule(l){3-8} 
\multicolumn{1}{c}{} &
  \multicolumn{1}{c}{} &
  mIoU$_{\text{full}}\uparrow$ &
  mIoU$_{\text{occ}}\uparrow$ &
  mIoU$_{\text{full}}\uparrow$ &
  mIoU$_{\text{occ}}\uparrow$ &
  mIoU$_{\text{full}}\uparrow$ &
  mIoU$_{\text{occ}}\uparrow$ \\ \hline 
BCNet~\cite{ke2021deep}            & CVPR 2021 & 15.10 &   -    & 74.90 &   -    & 44.00 &  -     \\
SLN~\cite{zhang2019learning}             & MM 2019   & 31.50 &    -   & 31.20 &     -  & 14.20 &    -   \\
ORCNN~\cite{follmann2019learning}           & WACV 2019 & 57.60 &   -    & 74.10 &   -    & 55.10 & -      \\
Mask-RCNN~\cite{he2017mask}       & ICCV 2017 & 63.85 &    -   & 74.62 &    -   & 60.13 &  -     \\
A3D~\cite{li20222d}             & ECCV 2022 & 64.27 &    -   & 74.71 &    -   & 61.43 &   -    \\
GIN~\cite{li2023gin}             & TMM 2024  & 72.46 &    -   & 78.23 &   -    & 68.31 &  -     \\
AISFormer~\cite{tran2022aisformer}       & BMVC 2022 & 72.69 & 13.75 & 86.81 & 30.01 & 81.53 & 48.54 \\
SAM$\dag$~\cite{kirillov2023segment}             & ICCV 2023 & 73.10 &  -     & 84.65 &   -    & 75.88 &    -   \\
SDXL-Inpainting$\dag$~\cite{podell2023sdxl} & ICLR 2024 & 73.65 &    -   & 80.53 &   -    & 76.19 &  -     \\
PCNet~\cite{zhan2020self}           & CVPR 2020 & 76.91 & 20.34 & 80.45 & 28.56 & 78.02 & 38.14 \\
VRSP~\cite{xiao2021amodal}            & AAAI 2021 & 78.98 & 22.92 & 88.08 & 35.17 & 80.70 & 47.33 \\
pix2gestalt$\dag$~\cite{ozguroglu2024pix2gestalt}     & CVPR 2024 & 79.08 &    -   & 81.82 &    -   & 81.45 &   -    \\
SDAmodal~\cite{zhan2024amodal}        & CVPR 2024 & 80.01 &  -     &   -    &    -   &    -   &   -    \\
C2F-Seg~\cite{gao2023coarse}         & ICCV 2023 & 80.28 &    -   & 89.10 &    -   & 82.22 &  -     \\
SAMBA$\dag$~\cite{liu2025towards}           & CVPR 2025 & 81.82 & -     & 90.87 &   -    & 88.47 &  -    \\ 
\rowcolor[HTML]{EFEFEF} 
\textbf{ShapeMoE (SG)}        & -         &  \underline{89.53}     &  \underline{30.68}     &  \underline{92.40}     &  \underline{37.95}     &   \underline{89.82}    &   \underline{49.70}    \\
\rowcolor[HTML]{EFEFEF}
\textbf{ShapeMoE (GMM)}         & -         &  \textbf{91.47}     &  \textbf{32.14}     &  \textbf{92.95}     &  \textbf{39.30}     &   \textbf{90.64}    &   \textbf{51.28}    \\ \bottomrule
\end{tabular}
\label{tab:comparison_with_sota}

\par
\noindent
\begin{minipage}[t]{\columnwidth}

\centering
\caption{Total expert count in ShapeMoE (SG)}
\singlecoltable{ccccc}{10}{
\toprule
Index & Total Experts & COCOA-cls      & D2SA           & KINS           \\ \midrule
1     & 1                    & 89.19          & 91.55          & 88.15          \\
2     & 2                    & 89.37          & 91.58          & 89.02          \\
3     & 4                    & \textbf{89.53} & \textbf{92.40} & \textbf{89.82} \\
4     & 8                    & 89.37          & 90.18          & 88.66          \\
5     & 16                   & 89.36          & 91.42          & 89.11          \\ \bottomrule
}
\label{tab:ab_expert_number}
\end{minipage}\hfill
\begin{minipage}[t]{\columnwidth}

\centering
\caption{Selected expert count in ShapeMoE (SG)}
\singlecoltable{ccccc}{10}{
\toprule
Index & Selected Expert Number & COCOA-cls      & D2SA           & KINS           \\ \midrule
1     & 1                      & \textbf{89.53} & \textbf{92.40} & \textbf{89.82} \\
2     & 2                      & 89.46          & 91.68          & 89.68 \\
3     & 3                      & 89.34          & 91.47          & 89.70          \\
4     & 4                      & 89.50          & 91.39          & 89.59          \\ \bottomrule
}
\label{tab:ab_selected_expert_number}
\end{minipage}
\end{table*}

\subsection{Datasets}
We evaluate ShapeMoE and the baselines on three amodal segmentation benchmarks.
(1) D2SA~\cite{follmann2019learning} extends D2S~\cite{follmann2018mvtec} with amodal masks for merchandise on a rotatable platform under varied poses and lighting. It contains 5,600 images and 28,720 instances across 60 categories.
(2) COCOA-cls~\cite{follmann2019learning}, based on COCO~\cite{lin2014microsoft}, covers diverse indoor and outdoor scenes. It contains 3,501 images and 10,592 annotated objects across 80 categories, with both visible and amodal masks.
(3) KINS~\cite{qi2019amodal} derives from KITTI~\cite{geiger2012we} and covers street scenes with pedestrians and vehicles. It provides 7,474 training images and 7,517 test images.

To ensure fair comparison, methods trained on the target datasets use the official training sets. All methods are evaluated on the validation sets of COCOA-cls and D2SA and the test set of KINS. Zero-shot methods follow the training settings in their original papers and use the same evaluation protocol as ours. Ground-truth amodal masks are used for training supervision where applicable and for evaluation. Regarding the input visible mask, we follow SAMBA by employing SAM1-predicted masks rather than the ground truth.

\subsection{Implementation Details}

We implement ShapeMoE in PyTorch~\cite{paszke2019pytorch} and train it with AdamW~\cite{loshchilov2017decoupled}, batch size 1, on an NVIDIA RTX 5090 (32 GB), without data augmentation. The Image Feature Encoder with an adapter~\cite{chen2024sam2} uses pretrained ViT-H~\cite{dosovitskiy2020image} weights. The Mask Embedding Encoder and segmentation experts use SAM2~\cite{ravi2024sam} weights, with remaining parameters randomly initialized~\cite{he2015delving}.

The GMM extension uses $T=0.1$, $\epsilon_{\mathrm{SK}}=0.05$, and unit weight for $\mathcal{L}_{\mathrm{bal}}^{\mathrm{GMM}}$. Its queue stores 3,840 detached latent vectors and starts filling in epoch 5. Training uses only the segmentation loss until the queue is full, then enables the balancing loss. Assignments use the previous queue plus the current sample ($B_{\mathrm{eff}}=3841$). Sinkhorn stops when the maximum absolute row-sum error and maximum relative column-sum error are both at most $10^{-3}$, targeting sums of $1$ and $B_{\mathrm{eff}}/N$, respectively.

We report mean intersection over union for complete masks (mIoU$_{\text{full}}$) and occluded regions (mIoU$_{\text{occ}}$)~\cite{gao2023coarse}, averaging results over three runs.

\subsection{Comparison with State-of-the-art Methods}
We compare ShapeMoE with state-of-the-art methods on the COCOA-cls and D2SA validation sets and the KINS test set, using SAM1-predicted visible masks following SAMBA. Methods marked with $\dag$ are zero-shot approaches pretrained on approximately 837,000 synthetic images with amodal annotations~\cite{ozguroglu2024pix2gestalt}.

\textbf{Quantitative Results.} 
As shown in Table~\ref{tab:comparison_with_sota}, we present the quantitative comparison with state-of-the-art methods. 
The GMM extension achieves the highest mIoU$_{\text{full}}$ and mIoU$_{\text{occ}}$ on all three datasets, followed by the single-Gaussian ShapeMoE. Compared with the single-Gaussian version, the GMM extension achieves gains of 1.94, 0.55, and 0.82 mIoU$_{\text{full}}$ and 1.46, 1.35, and 1.58 mIoU$_{\text{occ}}$ on COCOA-cls, D2SA, and KINS, respectively. 
Consistent with SAMBA, we adopt visible masks predicted by SAM1 instead of relying on ground-truth annotations to ensure fairness.
Compared to C2F-Seg, the state-of-the-art fully supervised method trained on the same datasets, ShapeMoE (SG) surpasses it by 9.25, 3.3, and 7.6 mIoU$_{\text{full}}$ for COCOA-cls, D2SA, and KINS, respectively. Furthermore, compared to SAMBA, the state-of-the-art zero-shot method, ShapeMoE (SG) achieves improvements of 7.71, 1.53, and 1.35 mIoU$_{\text{full}}$ across the three datasets. This performance gain can be attributed to ShapeMoE's explicit modeling of shape distributions and dynamic instance routing to specialized experts, enabling more accurate and shape-aware amodal mask prediction with increased model capacity.

\begin{figure*}[ht]
    \centering
    \includegraphics[width=0.95\linewidth]{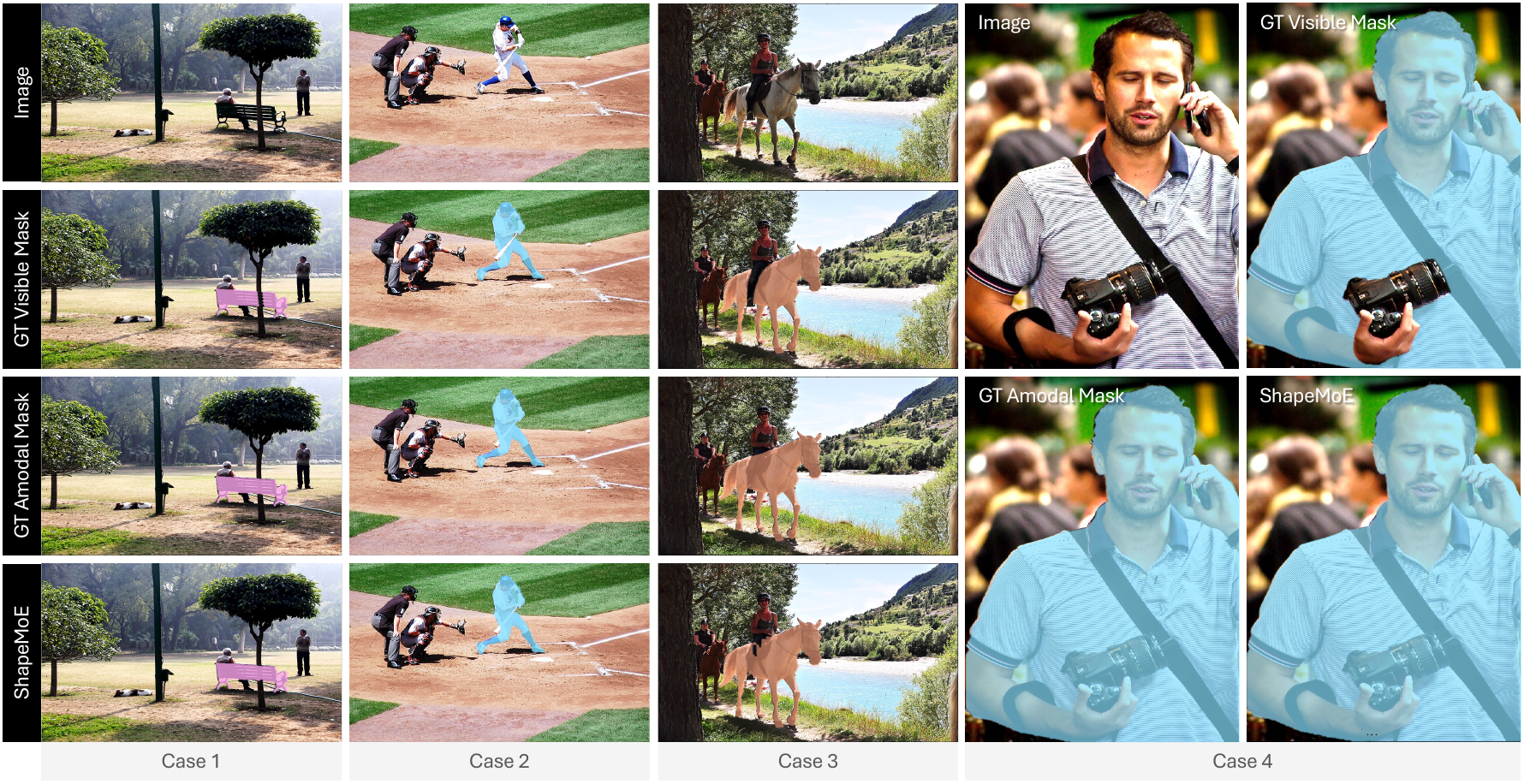}
    \caption{Qualitative results of the proposed ShapeMoE. Four representative cases are shown across various object categories, including bench, human, and horse, demonstrating ShapeMoE’s ability to handle complex occlusions and varied amodal shapes. Best viewed in color and zoomed in for details.}
    \label{fig:qualitative_results}
\end{figure*}

\textbf{Qualitative Results.} 
Fig.~\ref{fig:qualitative_results} presents qualitative results of the proposed ShapeMoE method on the COCOA-cls dataset. Four representative cases are shown from left to right. ShapeMoE demonstrates strong performance across various occlusion scenarios. For example, in Case 1, the rigid object bench is accurately predicted with a complete amodal mask, despite being partially occluded by the trunk of a tree. In Cases 2 and 4, the human subjects are partially occluded by a baseball bat crossing the body and a camera located within the body region, respectively. ShapeMoE successfully predicts complete amodal masks in both scenarios. In Case 3, although the horse is partially occluded by a human leg, ShapeMoE successfully handles the occlusion through the use of shape-specialized experts tailored for complex occlusions, producing a high-quality amodal mask.

\subsection{Ablation Study}
Ablation studies use ShapeMoE (SG), with the GMM extension evaluated separately.

\textbf{Effect of the Total Number of Experts.}
With one expert selected per sample, increasing the total number from 1 to 4 consistently improves performance across all datasets (Table~\ref{tab:ab_expert_number}), suggesting better modeling of shape diversity. Performance peaks at 4 experts but declines with 8 or 16, possibly due to under-utilization and harder routing optimization, showing a moderate expert count is sufficient to balance specialization and efficiency.

\begin{table}[!t]
\centering
\caption{Expert balancing loss in ShapeMoE (SG)}
\singlecoltable{ccccc}{10}{
\toprule
Index & Expert Balancing Loss & COCOA-cls      & D2SA           & KINS           \\ \midrule
1     &                       & 87.61          & 91.55          & 88.06          \\
2     & \checkmark            & \textbf{89.53} & \textbf{92.40} & \textbf{89.82} \\ \bottomrule
}
\label{tab:ab_expert_balancing_loss}
\end{table}

\textbf{Effect of Selected Expert Number.}
We vary the number of selected experts while fixing the total at 4. Selecting one expert per sample achieves the best overall performance, particularly on COCOA-cls and D2SA (Table~\ref{tab:ab_selected_expert_number}). Performance declines slightly with more selected experts, supporting sparse routing to encourage specialization and reduce computational redundancy.

\textbf{Effect of Expert Balancing Loss.}
We study expert balancing loss, designed to encourage uniform expert utilization during training. As shown in Table~\ref{tab:ab_expert_balancing_loss}, removing the balancing loss leads to consistent performance drops across all three datasets. These results indicate that the auxiliary loss improves segmentation performance.

\begin{table*}[!t]
\begin{minipage}[t]{\columnwidth}

\centering
\caption{Comparison with supervised and zero-shot methods}
\singlecoltable{cccc}{8}{
\toprule
Methods  & COCOA-cls & D2SA  & KINS  \\ \midrule
C2F-Seg  & 80.28     & 89.10  & 82.22 \\
SAMBA    & 81.82     & 90.87 & 88.47 \\
ShapeMoE (SG) & \textbf{84.07}     & \textbf{91.52} & \textbf{89.39} \\ \bottomrule
}
\label{tab:ab_finetune_vs_zero_shot}
\par
\centering
\caption{ShapeMoE (SG) across occlusion levels}
\singlecoltable{lccccc}{12}{ \toprule
    Methods     & FG-0  & FG-1  & FG-2  & FG-3  & Mean  \\  \midrule
    SAM2 & 92.17 & 88.86 & 86.79 & 79.52 & 89.19 \\
    ShapeMoE (SG)    & \textbf{92.19} & \textbf{88.90} & \textbf{88.28} & \textbf{81.53} & \textbf{89.53} \\ \rowcolor{gray!15}
    Improvement & +0.02 & +0.04 & +1.49 & +2.01 & +0.34 \\ \bottomrule    
}
\label{tab:ab_shapemoe_occ_levels}
\end{minipage}\hfill
\begin{minipage}[t]{\columnwidth}

\centering
\caption{Comparison of ShapeMoE (SG) and ShapeMoE (GMM)}
\singlecoltable{lccc}{8}{ \toprule
    Methods        & COCOA-cls & D2SA  & KINS  \\ \midrule
    SAMBA          & 81.82     & 90.87 & 88.47 \\
    SAM2           & 89.19     & 90.07 & 86.22 \\
    ShapeMoE (SG)       & 89.53     & 92.4  & 89.82 \\ 
    ShapeMoE (GMM) & \textbf{91.47}     & \textbf{92.95} & \textbf{90.64} \\ \bottomrule
}
\label{tab:ab_gaussian_model}
\par
\centering
\caption{Efficiency of ShapeMoE (SG) compared with SAM2}
\singlecoltable{lcccc}{10}{ \toprule
    Methods  & Parameters (M) & GFLOPs & Latency (ms) & FPS     \\ \midrule
    SAM2     & 224.93     & 406.08 & 67.32        & 14.85   \\
    ShapeMoE (SG) & 227.07     & 406.08 & 68.74        & 14.55   \\ \rowcolor{gray!15}
    Delta    & +0.95\%    & 0      & +2.1\%      & -2.0\% \\ \bottomrule
}
\label{tab:efficiency_comparison}
\end{minipage}
\end{table*}

\textbf{Zero-Shot Generalization.}
Following the zero-shot setting of SAMBA~\cite{liu2025towards}, we fine-tune ShapeMoE (SG) on pix2gestalt~\cite{ozguroglu2024pix2gestalt}, which contains approximately 837,000 images with amodal annotations. We evaluate directly on COCOA-cls, D2SA, and KINS without further training on these datasets. C2F-Seg serves as a supervised reference. Table~\ref{tab:ab_finetune_vs_zero_shot} shows that ShapeMoE achieves the best performance with 2.25, 0.65, and 0.92 mIoU$_{\text{full}}$ improvements on COCOA-cls, D2SA, and KINS, demonstrating the effectiveness of our shape-specific MoE design.

\textbf{Effect of the ShapeMoE design.}
Table~\ref{tab:ab_shapemoe_occ_levels} compares single-expert SAM2 and ShapeMoE (SG), which share the same backbone and are trained and evaluated on the COCOA-cls training and validation sets, respectively. Following SAMBA, FG-0 to FG-3 denote 0\%, 1--20\%, 20--40\%, and 40--70\% occlusion, respectively. ShapeMoE achieves higher mIoU$_{\text{full}}$ at all levels, with the largest gains under severe occlusion. These results confirm that the observed improvements are driven by our proposed architecture, rather than merely inheriting the capabilities of the powerful backbone model.

\textbf{GMM Extension.}
Single-Gaussian sampling is retained, with four shared Gaussian prototypes guiding weighted expert fusion (Section~\ref{sec:gmm_extension}). It improves mIoU$_{\text{full}}$ over ShapeMoE (SG) and individually fine-tuned SAM2 across all datasets (Table~\ref{tab:ab_gaussian_model}).

\textbf{Efficiency of ShapeMoE.}
We evaluate the efficiency of ShapeMoE (SG) on the COCOA-cls dataset. Compared with SAM2, parameters increase by 0.95\% (224.93M to 227.07M), with both at 406.08 GFLOPs. Latency increases by 2.1\% to 68.74 ms, while FPS decreases by 2.0\% to 14.55, indicating little computational overhead. These results indicate that our design improves amodal segmentation performance with little computational overhead.

\section{Conclusion}
We present ShapeMoE, a shape-specific sparse mixture-of-experts framework for amodal instance segmentation. We use Gaussian shape embeddings to guide the Shape-Aware Sparse Router in selecting a lightweight expert for each object. Shape-dependent routing allocates model capacity to different geometric patterns while maintaining sparse and efficient single-expert inference. This specialization supports the recovery of occluded regions across rigid and deformable objects with diverse shapes and occlusion patterns. Experiments on three benchmarks demonstrate improved segmentation of complete masks and occluded regions over state-of-the-art methods. We also explore a GMM extension that matches latent shapes to shared Gaussian prototypes and combines four expert predictions, achieving further gains. Future work will investigate lightweight architectures to improve efficiency.

\bibliographystyle{IEEEtran}
\bibliography{references}
\end{document}